\documentclass[letterpaper]{article} 
\usepackage[preprint]{aaai2027}  
\usepackage[hyphens]{url}  
\usepackage{graphicx} 
\usepackage{natbib}  
\usepackage{caption} 
\usepackage{longtable}
\usepackage{algorithm}
\usepackage{algorithmic}
\usepackage{amsmath}

\usepackage{longtable}
\usepackage[table]{xcolor}
\usepackage{multirow}

\usepackage{newfloat}
\usepackage{listings}
\DeclareCaptionStyle{ruled}{labelfont=normalfont,labelsep=colon,strut=off} 
\floatstyle{ruled}
\newfloat{listing}{tb}{lst}{}
\floatname{listing}{Listing}

\usepackage{booktabs}

\title{Evolving Creatives with Clicks: CTR-Driven Advertising image Editing via Attribute Disentanglement and Preference Alignment}
\author{
    Xu Zhang\textsuperscript{\rm 1,2}\thanks{Work done while interning at JD.com.}\equalcontrib,
    Haohan Wang\textsuperscript{\rm 2}\equalcontrib,
    Zheng Zhang\textsuperscript{\rm 2}\corresponding,
    Yingjie Xia\textsuperscript{\rm 3},
    Wei Feng\textsuperscript{\rm 2},
    Honghe Zhu\textsuperscript{\rm 2}\\
    Jingjing Lv\textsuperscript{\rm 2},
    Changping Peng\textsuperscript{\rm 2},
    Xuheng Qin\textsuperscript{\rm 2},
    Zehua Zhang\textsuperscript{\rm 2},
    Zheng Luo\textsuperscript{\rm 2},
    Qicheng Li\textsuperscript{\rm 1}\corresponding
}

\affiliations{
    \textsuperscript{\rm 1}TMCC, College of Computer Science, Nankai University,
    \textsuperscript{\rm 2}JD.COM,
    \textsuperscript{\rm 3}VCIP, Computer Science, Nankai University\\[0.5em]

    \textit{xu\_zhang@mail.nankai.edu.cn}
}

\begin{document}

\maketitle

\begin{abstract}
The optimization objective for e-commerce advertising creatives has shifted from aesthetic quality to online business metrics like click-through rate (CTR). Recent methods generate appealing ad images from product images by constructing CTR-based reward signals and refining prompt generation via Reinforcement Learning. Such paradigms equally penalize all attributes within an underperforming image, even though some are actually beneficial to clicks. In this work, we propose Evolving Creatives with Clicks (ECC), which reformulates creative optimization as fine-grained editing rather than generation from scratch. ECC selectively refines attributes that hurt CTR while preserving those that contribute positively. Specifically, faced with a rich attribute space across massive images, ECC first identifies 28 attributes significantly correlated with CTR via ANOVA and Random Forest regression. Based on attribute contrasts between images with disparate CTRs, we instruction-tune a prompt model to generate editing suggestions, then refine it with real click preferences since reference-derived suggestions are not uniformly beneficial. Furthermore, we introduce Consistent Preference Optimization (CPO) as ECC's preference optimization strategy, which enforces consistency across semantically equivalent shuffled suggestions to mitigate DPO overfitting and improve preference generalization. Extensive experiments demonstrate that ECC significantly improves the CTR of ad images, outperforming state-of-the-art models by a substantial margin. The complete data pipeline, including attribute extraction and analysis, the resulting dataset, and source code will be released upon acceptance.

\end{abstract}

\section{Introduction}

Advertising creatives, especially advertising images (ad images), serve as a key lever to capture user clicks~\cite{mishra2020learning,wang2021hybrid,chen2021automated}. 
Early works primarily focused on enhancing the aesthetics or controllability of generated ad images~\cite{li2023relation,du2024towards,chen2024virtualmodel,chen2501paid}. 
However, such improvements rarely translate into click-through rate (CTR) gains, as the criteria correlate loosely with online business performance.

Recent advances directly adopt CTR as the optimization objective~\cite{yang2024new,chen2025ctr}. 
The most prominent paradigm leverages CTR-derived rewards to fine-tune a prompt model via Direct Preference Optimization (DPO)~\cite{rafailov2023direct}, thereby generating prompts that yield better-performing images.
Despite achieving certain CTR improvements, this paradigm has three major limitations.
\begin{figure}[t]
\centering
\includegraphics[width=1.0\columnwidth]{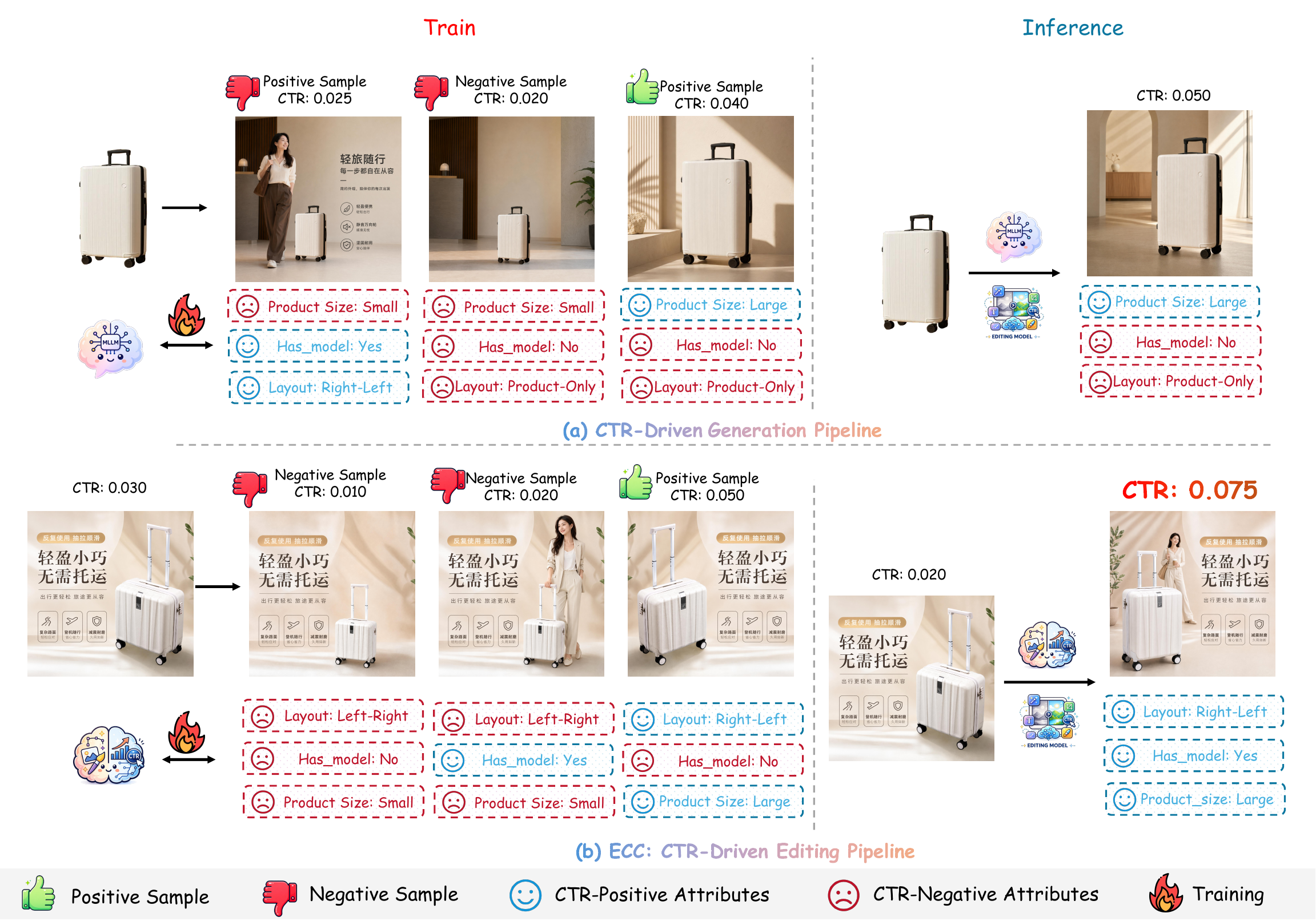} 
\caption{Comparison of Existing Advertising Creative Optimization Paradigms.}
\label{compare}
\end{figure}
First, as shown in Fig.~\ref{compare}(a), DPO treats prompts describing the entire ad image as positive or negative samples, causing all visual attributes in an underperforming image to be penalized equally, even though some may contribute positively to CTR.
Such erroneous signals are difficult to recover because promising visual ideas may appear only sparsely in the vast design space.
Second, ad images in practice originate from diverse production pipelines, while a unified framework that generates prompts and renders images from product images can hardly preserve or reproduce existing high-quality creatives.
Finally, DPO is prone to overfitting when preference supervision is limited~\cite{lin2024limited,chowdhury2024provably}.
Although expanding the preference dataset can alleviate this issue, collecting additional samples from real online feedback requires substantial image generation, traffic, and experimentation time, making overfitting reduction and data acquisition cost difficult to balance.

To address the first two limitations, we propose Evolving Creatives with Clicks (ECC), which reformulates creative optimization as fine-grained editing of existing ad images rather than generation from scratch.
As shown in Fig.~\ref{compare}(b), ECC selectively refines attributes that hurt CTR while preserving those that contribute positively.
We define 51 visual attribute dimensions based on large-scale e-commerce data and expert knowledge, and provide candidate values for most attributes to reduce ambiguity and avoid overly dispersed descriptions.
We then leverage MLLMs to extract attribute values from online ad images and identify 28 CTR-sensitive attributes through statistical analysis.
These attributes transform the vast editing space into a structured set of actionable dimensions, enabling more targeted CTR optimization.

Based on the selected attributes, ECC adopts a two-stage learning strategy.
In the first stage, we construct image pairs with disparate CTRs and derive attribute-level editing schemes. Since high-CTR creatives may still contain suboptimal attributes, the resulting suggestions are not always beneficial. We therefore randomly sample a subset of operations for each training instance to train the prompt model to generate diverse executable editing suggestions.
In the second stage, the instruction-tuned prompt model generates multiple suggestions for existing ad images, which are executed to produce new creatives. Since accurately predicting CTR from images remains challenging, with prior CTR reward models achieving only about 58\% accuracy~\cite{chen2025ctr}, we deploy these creatives online rather than relying on a learned reward model. Preference pairs are then constructed from real exposure and click feedback to distinguish effective suggestions and further refine the prompt model toward genuine CTR gains.

To mitigate DPO overfitting without further expanding the preference dataset, we propose Consistent Preference Optimization (CPO) as the preference optimization strategy of ECC.
Inspired by invariance learning in self-supervised learning~\cite{chen2020simple}, we identify that shuffling independent operations does not affect the edited image, so the original and shuffled chosen editing suggestions should receive consistent preferences.
CPO therefore introduces a lightweight consistency objective into DPO, encouraging the prompt model to focus on editing semantics rather than superficial ordering patterns and improving generalization without additional online data. Overall, our contributions are as follows:

\begin{itemize}

\item \textbf{CTR-Aware Attribute Mining and Dataset Construction Pipeline.}
We develop a unified pipeline for CTR-aware attribute mining and editing-suggestion data construction. By extracting 51 visual attributes from over 550,000 online ad images and identifying 28 CTR-sensitive dimensions through MLLM-based extraction and statistical analysis, we construct the first large-scale CTR-driven editing-suggestion datasets for instruction tuning and reinforcement learning.

\item \textbf{ECC: The First CTR-Driven Advertising Image Editing Framework.}
We propose ECC, the first framework to formulate ad image editing from a CTR-oriented perspective. Guided by CTR-sensitive attributes, it learns executable editing suggestions through instruction fine-tuning and aligns them with online click preferences through reinforcement learning.

\item \textbf{Consistent Preference Optimization Strategy.}
We propose Consistent Preference Optimization (CPO), which augments DPO with consistency regularization between chosen responses and their semantics-preserving shuffled variants, addressing the overfitting problem in preference learning. Large-scale online experiments show that ECC achieves the highest CTR gain among strong MLLM-based baselines.

\end{itemize}

\section{Related Work}

\subsection{E-commerce ad image Optimization}

Prior work on ad image optimization primarily focuses on visual quality and controllability. Methods such as AutoCO~\cite{chen2021automated}, PAID~\cite{chen2501paid}, retrieval-assisted advertising image generation~\cite{ku2023staging}, and instruction-guided editing models~\cite{brooks2023instructpix2pix,zhang2023magicbrush,huang2024smartedit,hui2024hq} improve visual appeal, composition, product presentation, or editing flexibility, but do not explicitly model user behavior.
CTR-driven methods leverage online interaction signals for creative refinement and ranking~\cite{mishra2020learning,wang2021hybrid,lin2022joint} and generation~\cite{wei2022creater,yang2024new,chen2025ctr}.
Ranking methods only select existing creatives, while generation-based methods often reuse fixed product foregrounds, limiting foreground diversity and creative variation. ECC instead edits existing creatives using real CTR feedback, enabling more flexible and diverse optimization.

\subsection{CTR Attribution and Visual Factor Analysis}

CTR prediction methods such as DeepFM~\cite{guo2017deepfm} and xDeepFM~\cite{lian2018xdeepfm} model feature interactions, while later studies incorporate visual information to improve prediction accuracy~\cite{ge2018image}. However, these models often lack interpretability and provide limited insight into which visual factors drive CTR differences.

Visual factor analysis commonly relies on statistical significance and feature importance. ANOVA~\cite{fisher1970statistical} evaluates whether different attribute values lead to significant CTR differences, while Random Forest regression~\cite{breiman2001random} captures nonlinear relationships and estimates attribute importance. Building on these methods, we identify CTR-sensitive visual attributes and use them to guide editing-oriented data construction.

\begin{figure*}[t]
\centering
\includegraphics[width=1.0\textwidth]{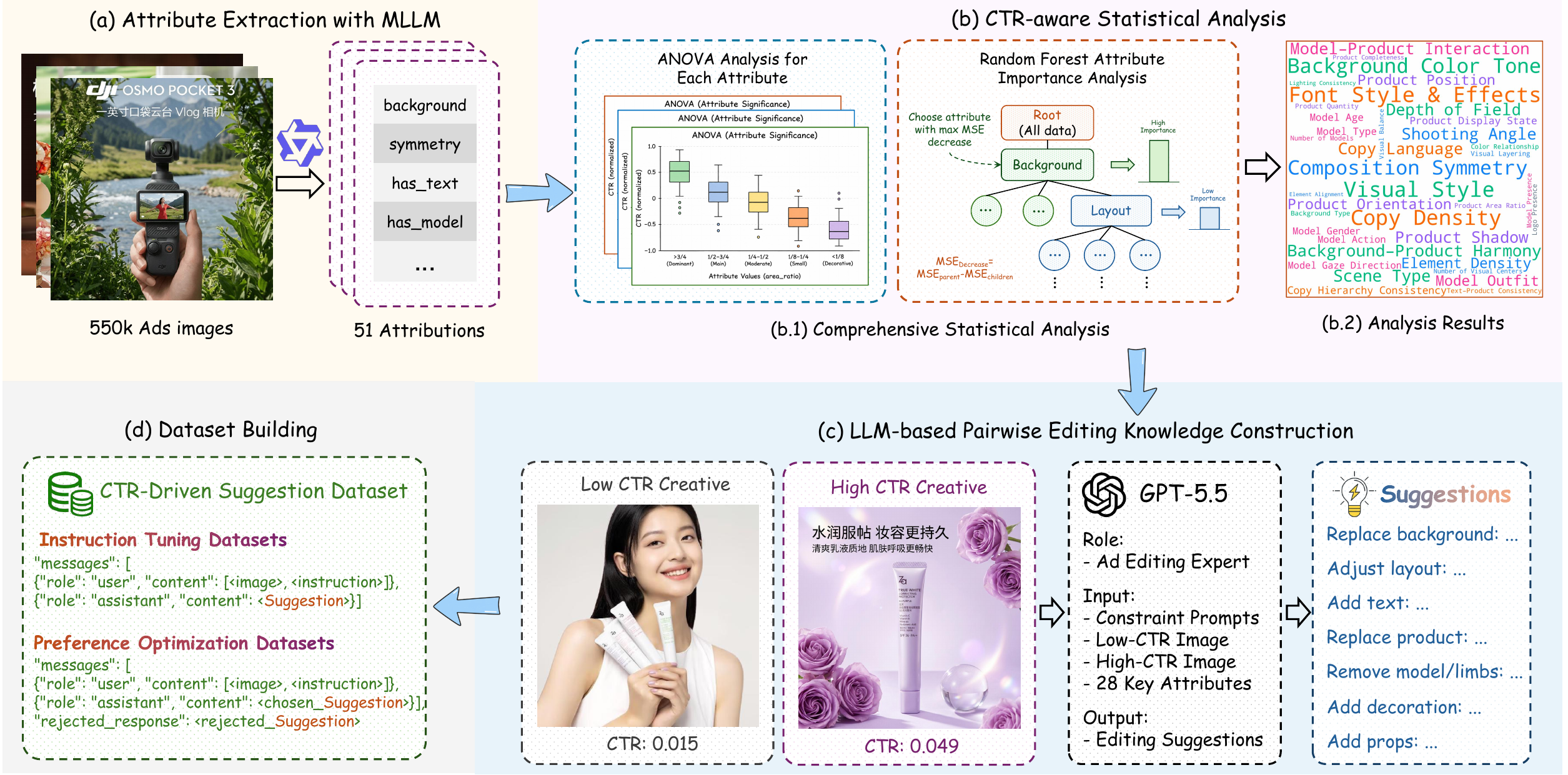} 
\caption{CTR-aware visual attribute mining and dataset construction pipeline.  (a) shows the extraction of 51 predefined visual attributes from over 550K ad images using Qwen3-VL-32B. (b) shows the joint ANOVA and Random Forest analysis for selecting 28 CTR-sensitive attributes. (c) shows executable editing suggestions generated by comparing same-product creatives with different CTRs under attribute-guided prompts. (d) shows the structures of the instruction-tuning and preference datasets.}
\label{dataset}
\end{figure*}

\subsection{Multimodal Instruction-Guided Image Editing}

MLLMs have shown strong visual understanding and reasoning capabilities. Recent MLLMs, including LLaVA~\cite{liu2023visual}, Qwen2-VL~\cite{Qwen2-VL}, Qwen2.5-VL~\cite{Qwen2.5-VL}, GPT-5.5~\cite{openai2026gpt55}, and Gemini 3.1 Pro Preview~\cite{google2026gemini31pro}, have substantially improved visual grounding and fine-grained image understanding.

Instruction-guided editing methods~\cite{brooks2023instructpix2pix,zhang2023magicbrush,huang2024smartedit,cao2024instruction,hui2024hq,sheynin2024emu,yu2025anyedit,xu2025insightedit,qian2026pico} enable flexible image modification, but mainly target generic visual edits without modeling CTR. ECC instead learns CTR-oriented editing suggestions from real creatives and online click feedback.

\begin{figure*}[t]
\centering
\includegraphics[width=1.0\textwidth]{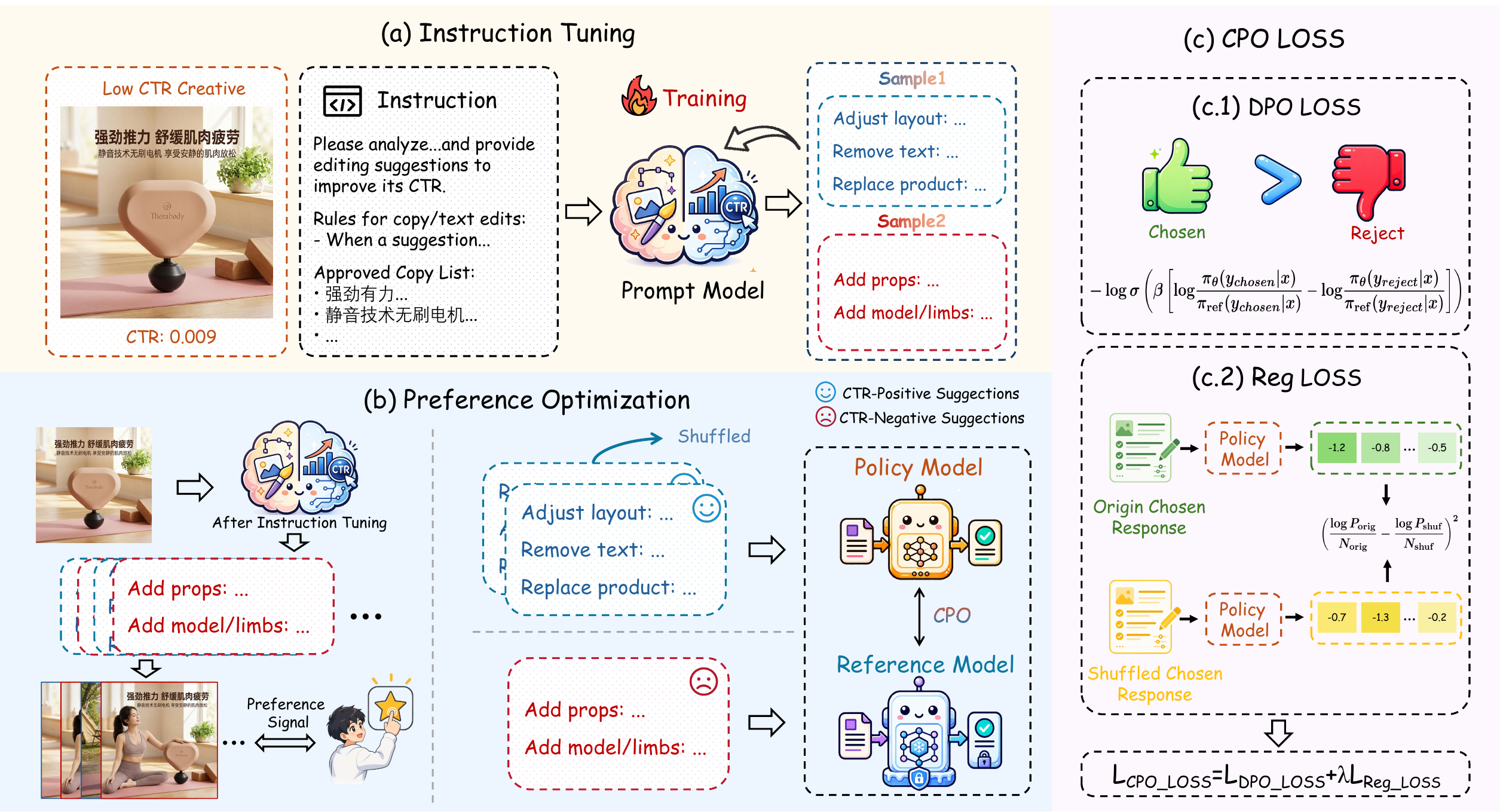} 
\caption{Training framework of ECC. (a) shows instruction tuning, where the model learns to generate CTR-driven editing suggestions. (b) shows preference optimization using original chosen, shuffled chosen, and rejected responses with policy and reference models. (c) presents the proposed CPO strategy, which combines the standard DPO loss with a consistency regularizer to enforce consistent preferences between chosen responses and their semantics-preserving shuffled variants.}
\label{method}
\end{figure*}

\section{Methods}

Our workflow begins by extracting visual attributes from large-scale online ad images and identifying CTR-sensitive dimensions through statistical analysis, which are then used to construct instruction-tuning data. The instruction-tuned model generates executable editing suggestions that are applied to source images for online deployment, and the resulting exposure and click feedback are used to construct preference data. Finally, CPO introduces consistency regularization over semantics-preserving shuffles to explicitly learn order-invariant preferences and mitigate overfitting.

\subsection{CTR-Aware Visual Attribute Mining}

As shown in Fig.~\ref{dataset}(a), we collect over 550K online ad images and use Qwen3-VL-32B~\cite{Qwen3-VL} to extract 51 predefined visual attributes, including composition and copywriting, among others. Most attributes use constrained candidate values for consistent extraction, while Scene Type and Visual Style permit open-ended descriptions. We then analyze the extracted attribute values and their distributions, pre-excluding 14 attributes with unstable extraction, subjective or redundant semantics, or weak attribution signals, and retain 37 candidates for statistical analysis.

To identify visual attributes associated with click-through rate (CTR), we combine \textbf{ANOVA} significance with \textbf{Random Forest} feature importance, as illustrated in Fig.~\ref{dataset}(b.1). To reduce product-level differences in popularity and user demand, CTR is first normalized within each product. For ANOVA~\cite{fisher1970statistical}, samples are grouped by their extracted attribute values, and the between-group and within-group variations in normalized CTR are compared. The resulting $F$-statistic is converted into a $p$-value under the corresponding $F$ distribution, where a smaller value indicates a stronger association with CTR. Since ANOVA evaluates each attribute independently, we further train a Random Forest regressor~\cite{breiman2001random} to predict normalized CTR and estimate attribute contributions under multi-attribute interactions. Categorical attributes are converted into one-hot features, and the contribution of each split is measured by the weighted reduction in mean squared error:
{\small
\begin{equation}
\Delta\mathrm{MSE}
=
\mathrm{MSE}(\mathcal{D})
-
\frac{|\mathcal{D}_L|}{|\mathcal{D}|}\mathrm{MSE}(\mathcal{D}_L)
-
\frac{|\mathcal{D}_R|}{|\mathcal{D}|}\mathrm{MSE}(\mathcal{D}_R),
\label{eq:mse_reduction}
\end{equation}
}
where $\mathcal{D}_L$ and $\mathcal{D}_R$ denote the subsets produced by the split. Importance scores of one-hot branches for each attribute are summed and normalized as attribute-level contributions.

The final selection considers ANOVA significance, Random Forest contribution, distribution reliability, semantic redundancy, and practical editability. 
Most attributes follow the two quantitative signals, with a few boundary cases adjusted accordingly.
For example, Product Quantity is retained for its relatively high Random Forest contribution and direct editability, while Copy Hierarchy Consistency and Product Shadow are excluded due to redundancy or unstable distributions. This yields 28 CTR-sensitive attributes from 37 candidates. Their relative importance is visualized in Fig.~\ref{dataset}(b.2), where larger words indicate higher importance; complete results are provided in the supplementary material.

\subsection{Instruction-Guided Editing Learning}

As shown in Fig.~\ref{method}(a), we perform instruction tuning using the CTR-driven editing-suggestion dataset constructed based on the 28 selected attributes. Given a source ad image $I_r$ and an instruction $q$, the prompt model learns to generate executable editing suggestions for visual factors such as composition, copywriting, and background.

Since the attribute-guided supervision provides only an initial editing signal and not every operation necessarily improves CTR, we introduce Stochastic Editing Operation Sampling. For each training instance, 1--3 operations are randomly sampled from the complete suggestion and concatenated as the target response. This exposes the model to diverse operation combinations while leaving subsequent online preference optimization to identify modifications that genuinely improve CTR.

Let $\tilde{\mathbf{y}}=(\tilde{y}_1,\ldots,\tilde{y}_T)$ denote the sampled response. The model is optimized using the standard autoregressive loss:
\begin{equation}
\mathcal{L}_{\mathrm{SFT}}
=
-
\sum_{t=1}^{T}
\log P_{\theta}
\left(
\tilde{y}_t
\mid
I_r,q,\tilde{y}_{<t}
\right),
\label{eq:loss}
\end{equation}
where $\tilde{y}_t$ denotes the $t$-th token, $\tilde{y}_{<t}$ denotes the preceding tokens, and $T$ is the response length. After instruction tuning, the prompt model acquires the initial capability to generate executable CTR-oriented editing suggestions, which are further refined using real online click feedback.

\subsection{Preference Optimization}

As shown in Fig.~\ref{method}(b), the instruction-tuned prompt model generates multiple editing suggestions for each source image. These suggestions are executed by the image-editing backend, and the resulting creatives are deployed online. Their real exposure and click feedback are then used to construct preference pairs, allowing the model to learn which editing operations are more likely to improve CTR.

Given an input $X$, let $Y^{+}$ and $Y^{-}$ denote suggestions with higher and lower post-editing CTR, respectively. We first adopt the DPO objective to align the model with these online click preferences:
{\small
\begin{equation}
\mathcal{L}_{\mathrm{DPO}}
=
-\log \sigma
\left(
\beta
\left[
\log\frac{\pi_{\theta}(Y^{+}|X)}
{\pi_{\mathrm{ref}}(Y^{+}|X)}
-
\log\frac{\pi_{\theta}(Y^{-}|X)}
{\pi_{\mathrm{ref}}(Y^{-}|X)}
\right]
\right),
\label{eq:dpo_loss}
\end{equation}
}
where $\pi_{\theta}$ and $\pi_{\mathrm{ref}}$ denote the policy and frozen reference models, respectively. This preference-learning stage further refines the initial attribute-guided supervision according to real online outcomes.

However, standard DPO may overfit to superficial patterns in editing suggestions rather than underlying editing semantics. Inspired by invariance learning in self-supervised learning~\cite{chen2020simple}, we identify a task-specific order invariance: reordering independent editing operations does not affect the final edited image, so the original and shuffled chosen suggestions should receive consistent preferences. We construct $\widetilde{Y}^{+}$ by randomly shuffling the independent operations in each chosen suggestion $Y^{+}$ while preserving its semantics and editing effect. As illustrated in Fig.~\ref{method}(c), the consistency term encourages them to receive similar length-normalized log-likelihoods~\cite{lu2024eliminating}:
\begin{equation}
\mathcal{L}_{\mathrm{Reg}}
=
\left(
\frac{\log\pi_{\theta}(Y^{+}|X)}{|Y^{+}|}
-
\frac{\log\pi_{\theta}(\widetilde{Y}^{+}|X)}
{|\widetilde{Y}^{+}|}
\right)^2,
\label{eq:consistency_loss}
\end{equation}
where $|Y|$ denotes the response length in tokens. The final objective is
\begin{equation}
\mathcal{L}_{\mathrm{CPO}}
=
\mathcal{L}_{\mathrm{DPO}}
+
\lambda\mathcal{L}_{\mathrm{Reg}},
\label{eq:cpo_loss}
\end{equation}
where $\lambda$ controls the regularization strength. CPO retains the DPO objective derived from online CTR feedback while explicitly enforcing preference consistency under semantics-preserving order variations.

\section{Datasets}

To support the two training stages, we construct an instruction-tuning dataset for learning executable editing suggestions and an online preference dataset for reinforcement optimization. The former provides the model with an initial capability to identify and describe actionable modifications, while the latter further aligns these suggestions with real user click preferences.

As shown in Fig.~\ref{dataset}(c), we first group sufficiently exposed creatives by product and construct image pairs with clear CTR differences. For each pair, the lower-CTR image is treated as the editing target, while the higher-CTR image serves as a reference that provides potentially beneficial visual cues. Guided by the 28 selected attributes, GPT-5.5~\cite{openai2026gpt55} compares the two creatives and generates executable editing suggestions covering visual factors such as composition, copywriting, product presentation, and background design. The reference image and extracted attribute values are used only during data construction and are not included in the model input. Each instruction-tuning sample therefore consists of the lower-CTR creative and an editing instruction as input, with the generated editing suggestion as the target response. After data filtering and quality control, this process produces 101,699 training samples.

After instruction tuning, the model generates multiple candidate editing suggestions for each source image. These suggestions are executed by the image-editing backend to produce corresponding creatives, which are then reviewed and deployed online to collect real exposure and click feedback. For the same source image, suggestions that lead to higher and lower post-editing CTRs are selected as chosen and rejected responses, respectively. Each preference sample contains the same source image and instruction as input, together with one chosen suggestion and one rejected suggestion, ensuring that the preference signal reflects differences in editing strategies rather than input conditions. This process produces 68,704 preference pairs. The structures of both datasets are shown in Fig.~\ref{dataset}(d).

\section{Experiments}

\subsection{Experimental Setup}

\subsubsection{Implementation Details.}
We use Qwen3-VL-8B-Instruct as the base multimodal model for generating editing suggestions. All models are trained using the SWIFT~\cite{zhao2024swiftascalablelightweightinfrastructure} framework on 4 NVIDIA A800 GPUs. For instruction-guided editing learning, we apply LoRA to all linear layers with rank 8 and alpha 32, and train the model for 3 epochs with a learning rate of $1\times10^{-4}$. For preference optimization, we initialize from the instruction-tuned adapter and optimize using the proposed CPO objective. This stage uses LoRA training for 4 epochs with a learning rate of $5\times10^{-5}$, and the regularization weight $\lambda$ is set to 1.

\subsubsection{Online Evaluation Protocol.}
To evaluate the practical effectiveness of the generated editing suggestions, we conduct a one-week online experiment on a large-scale e-commerce platform covering 44 product categories. For all compared methods, Gemini 3 Pro Image Preview~\cite{google2025gemini3proimage} applies the generated suggestions to the original images, and the resulting creatives are deployed online. Before deployment, all edited images are reviewed by professional practitioners to ensure visual quality and foreground--background compatibility. Their real exposure and click feedback are then used to determine the CTR effects of different editing suggestions. The experiment accumulates approximately 20 million impressions, providing reliable feedback for both preference construction and online evaluation.

\subsubsection{Evaluation Metrics.}
We evaluate each method using the aggregate click-through rate (CTR) gain between edited and original advertising creatives:
\begin{equation}
\Delta_{\mathrm{CTR}}
=
\frac{\mathrm{CTR}_{\mathrm{edit}}-\mathrm{CTR}_{\mathrm{orig}}}
{\mathrm{CTR}_{\mathrm{orig}}}
\times 100\%.
\end{equation}
Here, $\mathrm{CTR}_{\mathrm{edit}}$ and $\mathrm{CTR}_{\mathrm{orig}}$ denote the impression-weighted CTRs of the edited and original creatives, respectively. 

\subsection{Online Results}

\subsubsection{Main Online Results.}
Tab.~\ref{tab:main_online_results} reports the main online comparison in terms of CTR gain. ECC achieves the highest improvement of 6.86\%, substantially outperforming the existing CTR-oriented method CAIG~\cite{chen2025ctr}, which achieves only 0.20\%. ECC also surpasses GPT and Gemini by 3.61 and 1.34 percentage points, respectively. DPO$_{S1}$ and DPO$_{S2}$ indirectly exploit order invariance through sentence-order shuffling on one-third and all training samples, whereas CPO enforces it directly. Vanilla DPO and its shuffled-response variants yield only limited gains. Under the current traffic scale, ECC is expected to generate approximately RMB 117 million in additional annual revenue. These results demonstrate the effectiveness of ECC in generating CTR-oriented editing suggestions and validate the proposed preference optimization strategy.
\begin{table}[h]
\centering
\small
\setlength{\tabcolsep}{6pt}
\begin{tabular}{lc}
\toprule
\textbf{Method} & \textbf{CTR Gain} $\uparrow$ \\
\midrule
CAIG~\cite{chen2025ctr} & +0.20\% \\
DPO & +0.68\% \\
DPO$_{S1}$ & +1.12\% \\
DPO$_{S2}$ & +1.29\% \\ 
GPT & +3.25\% \\
Gemini & +5.52\% \\  
\textbf{ECC} & \textbf{+6.86\%} \\
\bottomrule
\end{tabular}
\caption{Main online results in terms of CTR gain.}
\label{tab:main_online_results}
\end{table}

\subsubsection{Analysis Across Product Categories and Advertising Scenarios.}
As shown in the \textit{Product Category} and \textit{Advertising Scenario} blocks of Tab.~\ref{tab:online_ctr_gain}, ECC consistently improves CTR across all evaluated categories and placements. The gains are particularly strong in Home Appliances, Computers, and Beauty, as well as on the Order Page and Product Detail Page. These results demonstrate that ECC generalizes across diverse products and traffic contexts rather than benefiting only specific categories or scenarios.

\begin{table}[h]
\centering
\small
\setlength{\tabcolsep}{6pt}
\begin{tabular}{lc}
\toprule
\textbf{Setting} & \textbf{CTR Gain} $\uparrow$ \\
\midrule  
\multicolumn{2}{l}{\textit{Product Category}} \\
Home Appliances & +14.01\% \\
Computers & +11.61\% \\
Beauty & +11.37\% \\
Digital Products & +3.92\% \\
Personal Care & +3.28\% \\
\midrule  
\multicolumn{2}{l}{\textit{Advertising Scenario}} \\
Order Page & +13.32\% \\
Product Detail Page & +12.79\% \\
Guess You Like & +5.53\% \\
Homepage Recommendation Feed & +5.23\% \\
Shopping Cart Page & +4.23\% \\
\midrule 
All & +6.86\% \\
\bottomrule
\end{tabular}
\caption{CTR gains across different product categories and advertising scenarios.}
\label{tab:online_ctr_gain}
\end{table}

\subsection{Aesthetic Evaluation}

We evaluate edited-image quality using LAION Aesthetic~\cite{schuhmann2022laion}, Aesthetic Predictor v2.5~\cite{zhai2023sigmoid,discus0434aesthetic}, PickScore~\cite{kirstain2023pick}, and GPT-5.5. For each image group, methods are ranked by each evaluator and averaged across groups. The overall score is the mean of the four evaluator-specific ranks, with lower values indicating better performance. As shown in Tab.~\ref{tab:aesthetic}, ECC achieves the best overall ranking, suggesting that CTR-oriented editing preserves visual quality. However, evaluator rankings are inconsistent and do not fully align with the online CTR gains in Tab.~\ref{tab:main_online_results}, indicating that aesthetic quality and advertising effectiveness are not equivalent.

\begin{table}[t]
\centering
\small
\setlength{\tabcolsep}{3.5pt}
\begin{tabular}{lccccc}
\toprule
\textbf{Method}  &
\textbf{LAION}  &
\textbf{Predictor}  &
\textbf{PickScore}  &
\textbf{VLM} &
\textbf{Avg.} $\downarrow$ \\
\midrule  
\textbf{ECC} & \textbf{2.47} & 2.41 & \textbf{2.36} & 2.97 & \textbf{2.55} \\
DPO$_{S2}$ & 2.53 & \textbf{2.34} & 2.62 & 3.01 & 2.63 \\
DPO$_{S1}$ & 2.61 & 2.45 & 2.64 & 2.92 & 2.66 \\
DPO & 3.34 & 2.59 & 3.20 & 2.98 & 3.03 \\
Gemini & 3.66 & 3.72 & 2.95 & \textbf{2.73} & 3.27 \\
GPT & 3.72 & 3.70 & 3.40 & 2.78 & 3.40 \\
Original & 2.68 & 3.79 & 3.82 & 3.62 & 3.48 \\
\bottomrule
\end{tabular}
\caption{Average rankings under four aesthetic evaluators.}
\label{tab:aesthetic}
\end{table}

\begin{figure*}[t]
\centering
\includegraphics[width=1.0\textwidth]{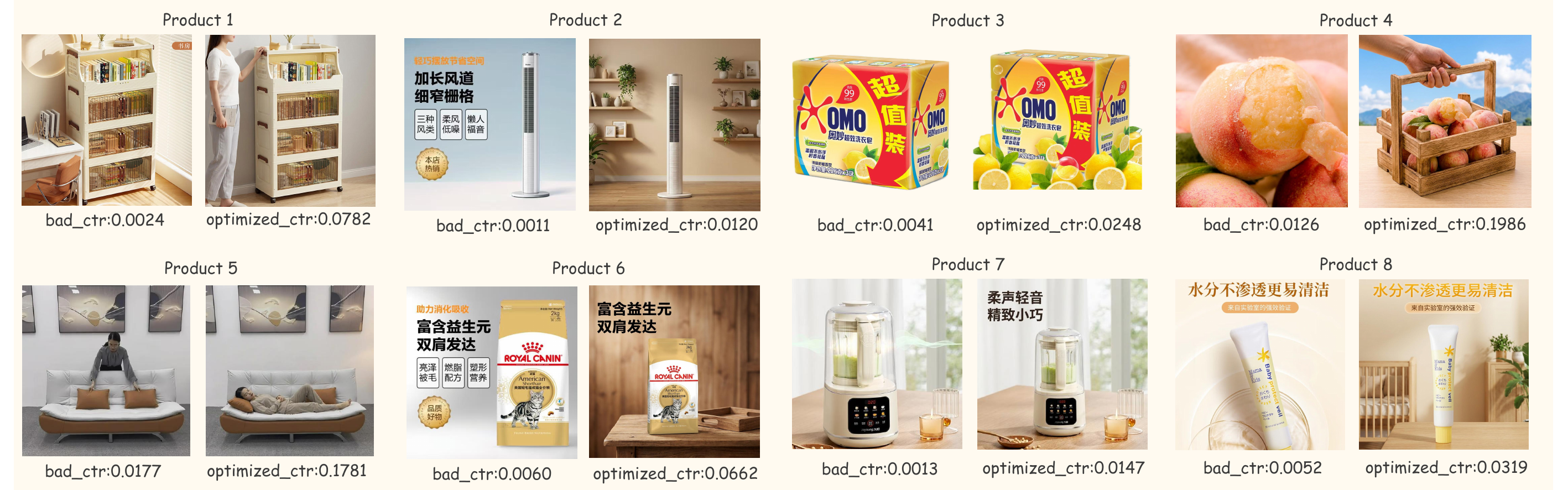}
\caption{Qualitative examples of CTR-driven edits from original images on the left to optimized images on the right.}
\label{fig:case_study}
\end{figure*}

\subsection{Order Robustness Analysis}

To evaluate CPO, we perform inference on approximately 1,000 editing samples using models trained with different preference optimization strategies. For each chosen suggestion, we construct a shuffled variant by permuting its editing operations without changing its content or editing effect.

Following the DPO implicit reward formulation, we compute length-normalized scores from the policy and reference models. We report three margins: $\Delta_{c\rightarrow r}$ between chosen and rejected responses, $\Delta_{c\rightarrow s}$ between chosen and shuffled chosen responses, and $\Delta_{s\rightarrow r}$ between shuffled chosen and rejected responses. Larger $\Delta_{c\rightarrow r}$ and $\Delta_{s\rightarrow r}$ indicate stronger preference discrimination, while smaller $\Delta_{c\rightarrow s}$ indicates better order consistency.

As shown in the \textit{Method Comparison} block of Tab.~\ref{tab:order_robustness}, DPO exhibits similar $\Delta_{c\rightarrow r}$ and $\Delta_{c\rightarrow s}$ values, with $\Delta_{s\rightarrow r}$ close to zero, indicating sensitivity to sentence-order variations. DPO$_{S2}$ improves order consistency through shuffled data augmentation, while ECC achieves the lowest $\Delta_{c\rightarrow s}$ and the highest $\Delta_{s\rightarrow r}$ while maintaining strong chosen-rejected discrimination. These results demonstrate that CPO more effectively learns order-invariant preferences.

\begin{table}[h]
\centering
\small
\setlength{\tabcolsep}{5pt}
\begin{tabular}{lccc}
\toprule
\textbf{Method} &
$\Delta_{c\rightarrow r}\uparrow$ &
$\Delta_{c\rightarrow s}\downarrow$ &
$\Delta_{s\rightarrow r}\uparrow$ \\
\midrule

\multicolumn{4}{l}{\textit{Method Comparison}} \\
DPO & 0.0213 & 0.0210 & 0.0003 \\
DPO$_{S2}$ & 0.0316 & 0.0126 & 0.0191 \\  
\textbf{ECC} & \textbf{0.0316} & \textbf{0.0119} & \textbf{0.0197} \\

\midrule
 
\multicolumn{4}{l}{\textit{Ablation on Regularization Weight}} \\
ECC ($\lambda=0.75$) & 0.0216 & 0.0121 & 0.0104 \\
ECC ($\lambda=1.25$) & 0.0225 & \textbf{0.0091} & 0.0034 \\  
\textbf{ECC ($\lambda=1.00$)} & \textbf{0.0316} & 0.0119 & \textbf{0.0197} \\

\bottomrule
\end{tabular}
\caption{Order robustness under DPO implicit rewards.}
\label{tab:order_robustness}
\end{table}

\subsection{Validation of CTR-Sensitive Visual Attributes}

To validate the discovered CTR-sensitive attributes, we formulate CTR estimation as a pairwise classification task: given two images of the same product, the model predicts which one has a higher CTR. We retain image pairs with sufficient exposure, resulting in 37,020 training pairs and 9,887 test pairs. Presenting each pair in both orders yields 74,040 training samples and 19,774 test samples.

We evaluate general-purpose MLLMs and a task-specific CTR prediction model under three input settings: image pairs only, image pairs with prompts describing all 51 predefined attributes, and image pairs with prompts describing the selected 28 attributes. As shown in Tab.~\ref{tab:attribute_validation}, the task-specific model consistently outperforms general-purpose MLLMs. Attribute prompts generally improve the image-only setting, while the selected 28 attributes achieve the highest accuracy of 59.56\%, indicating that they retain CTR-relevant information while excluding redundant or unstable factors.

\begin{table}[h]
\centering
\small
\setlength{\tabcolsep}{6pt}
\begin{tabular}{lccc}
\toprule
\multirow{2}{*}{\textbf{Model}} &
\multicolumn{3}{c}{\textbf{Input Setting}} \\
\cmidrule(lr){2-4}
&
\textbf{Image Only} &
\textbf{+51 Attr.} &
\textbf{+28 Attr.} \\
\midrule
GPT & 47.58 & 49.50 & 49.86 \\
Gemini & 48.93 & 48.24 & 49.44 \\
 
\textbf{Ours} &
\textbf{58.49} &
\textbf{59.46} &
\textbf{59.56} \\
\bottomrule
\end{tabular}
\caption{Pairwise CTR prediction accuracy (\%) under different input settings.}
\label{tab:attribute_validation}
\end{table}

\subsection{Ablation Study}
\subsubsection{Ablation on Preference Optimization.}
Tab.~\ref{tab:ablation_results} compares standard DPO, partial and full response-order shuffling augmentation, and the proposed Consistent Preference Optimization Strategy.
DPO brings only limited CTR improvement, indicating that directly fitting offline preference pairs is insufficient for learning robust editing preferences.
Although shuffled-response augmentation slightly improves CTR gain by exposing the model to different operation orders, it remains much weaker than CPO.
CPO explicitly constrains semantically equivalent editing suggestions to receive consistent preferences, helping the model focus on editing semantics rather than superficial ordering patterns.
This provides a more direct way to encode task-specific invariance than simply expanding the training data with shuffled variants.
As a result, ECC achieves the highest CTR gain of 6.86\%, validating the proposed preference optimization design.

\begin{table}[h]
\centering
\small
\setlength{\tabcolsep}{5pt}
\begin{tabular}{lc}
\toprule
\textbf{Variant} & \textbf{CTR Gain} \\
\midrule
DPO & +0.68\% \\
DPO$_{S1}$ & +1.12\% \\
DPO$_{S2}$ & +1.29\% \\  
\textbf{ECC w/ CPO} & \textbf{+6.86\%} \\
\bottomrule
\end{tabular}
\caption{Ablation results on preference optimization.}
\label{tab:ablation_results}
\end{table}

\subsubsection{Sensitivity Analysis of the Regularization Weight.}

To examine the sensitivity of CPO to the regularization weight $\lambda$, we evaluate ECC with several weight settings. As shown in the \textit{Ablation on Regularization Weight} block of Tab.~\ref{tab:order_robustness}, all tested settings improve order robustness over DPO, confirming that CPO remains effective across different weights and is not overly sensitive to a specific configuration. Among them, a weight of 1.00 provides the best overall trade-off between order invariance and preference discrimination, and is therefore adopted in the main experiments.

\subsection{Qualitative Analysis}

Fig.~\ref{fig:case_study} shows representative CTR-driven editing examples, with original and edited creatives shown on the left and right, respectively. The edited images feature cleaner layouts, more suitable backgrounds, clearer product presentation, and more focused copy, resulting in higher online CTR. Additional examples with corresponding editing suggestions are provided in the supplementary material.

\section{Conclusion}
In this paper, we propose ECC, a CTR-driven framework that improves existing ad images through fine-grained editing rather than generation from scratch. Using large-scale online data and statistical analysis, ECC identifies 28 CTR-sensitive attributes from 51 predefined dimensions and uses them to construct actionable editing supervision for targeted creative refinement. It first learns to generate executable editing suggestions through instruction tuning and is then further refined through preference optimization based on real click feedback. We further propose Consistent Preference Optimization (CPO) as ECC's preference optimization strategy, augmenting DPO with a consistency regularizer to mitigate overfitting to spurious ordering patterns and improve preference generalization. Large-scale online experiments show that ECC outperforms strong MLLM-based baselines and achieves the highest aggregate CTR gain across diverse product categories and advertising scenarios, demonstrating the effectiveness and practical value of fine-grained attribute editing for creative optimization.

\bibliography{aaai2027}

\clearpage
\onecolumn
\begin{center}
\LARGE \textbf{Evolving Creatives with Clicks: CTR-Driven Advertising image Editing via Attribute Disentanglement and Preference Alignment} \\ 
\vspace{1em}
\Large\textbf{Supplementary Material} \\
\end{center}
\vspace{2em}

\section{Overview}

This supplementary material provides additional details and qualitative results of ECC. We organize the supplementary material into three parts. First, we introduce the CTR-aware visual attribute mining process, which establishes a fine-grained attribute space for analyzing the relationship between advertising creatives and CTR. Second, we present representative instruction-tuning examples, illustrating how CTR-sensitive attributes are used to construct executable editing suggestions from same-product creative pairs. Third, we provide additional qualitative results to demonstrate the effectiveness and diversity of CTR-driven image editing across different products and editing scenarios.

\section{A. CTR-Aware Visual Attribute Mining}

This section provides additional details of the CTR-aware visual attribute mining process. We construct a visual attribute schema with 51 attributes for large-scale advertising creative analysis. These attributes are extracted by Qwen3-VL-32B and further analyzed through statistical evaluation and contribution modeling. After preliminary screening, 37 candidate attributes are evaluated using one-way ANOVA and Random Forest contribution. Considering statistical significance, attribute reliability, semantic redundancy, and practical editability, we finally identify 28 CTR-sensitive attributes.

\subsection{A.1 Visual Attribute Schema}

To systematically characterize the visual factors of advertising creatives, we construct a hierarchical attribute schema covering composition and layout, copywriting and text design, background and visual style, product presentation, model and interaction, and logo-related information. For each attribute, we define a candidate label space to reduce semantic fragmentation during large-scale extraction. Most attributes follow the \textit{Strict} setting, where Qwen3-VL-32B selects values from predefined candidates. Attributes with the \textit{Extensible} setting allow additional values when predefined candidates cannot sufficiently describe the visual content. The complete attribute schema and candidate values are presented in Tab.~\ref{tab:attribute_schema}.

\begin{center}
\small
\setlength{\tabcolsep}{5pt}
\renewcommand{\arraystretch}{1.08}

\begin{longtable}{p{0.24\textwidth}p{0.68\textwidth}}
\caption{Visual attribute schema.}
\label{tab:attribute_schema} \\

\toprule
\textbf{Attribute} & \textbf{Label Space / Candidate Values} \\
\midrule
\endfirsthead

\toprule
\textbf{Attribute} & \textbf{Label Space / Candidate Values} \\
\midrule
\endhead

\midrule
\multicolumn{2}{r}{\textit{Continued on next page}} \\
\endfoot

\bottomrule
\endlastfoot

\multicolumn{2}{l}{\textit{Composition and Layout}} \\
Image Aspect Ratio & Strict: 1:1, 16:9, 9:16, 3:4, 4:3, 3:5, 4:5. \\
Composition Symmetry & Strict: central symmetry, rule of thirds, horizontal symmetry, vertical symmetry, asymmetric balance, diagonal composition, radial composition. \\
Element Density & Strict: sparse, medium, dense. \\
Visual Balance & Extensible: balanced, cluttered, visually unbalanced, or other layout balance descriptions. \\
Element Alignment & Strict: left-aligned, centered, right-aligned, top-aligned, bottom-aligned. \\
Number of Visual Centers & Strict: no clear center, single center, double centers, multiple centers. \\
Visual Layering & Strict: single-layer flat layout, foreground-background layering, multi-layer composition. \\
Shooting Angle & Strict: eye-level, high angle, low angle, full view, flat lay, close-up, macro. \\

\midrule
 
\multicolumn{2}{l}{\textit{Copywriting and Text Design}} \\
Copy Content & Open-ended: promotional copy extracted from the image. \\
Copy Density & Numeric: number of promotional text blocks. \\
Copy Language & Strict: Chinese, English, mixed language. \\
Text--Product Consistency & Extensible: consistent with product, intentional contrast, neutral, severe mismatch, or other consistency descriptions. \\
Copy Alignment & Strict: left, right, center, justified, vertical, outline. \\
Copy Position & Strict: top-left, top-right, bottom-left, bottom-right, left-center, right-center, top-center, bottom-center, center. \\
Character Spacing & Strict: large, medium, small. \\
Font Size & Strict: extra large, large, medium, small, tiny. \\
Font Type & Extensible: minimalist, handwritten, retro, calligraphic, rounded, bold, artistic, or other font styles. \\
Font Style \& Effects & Extensible: no effect, gradient, outline, shadow, glow, metallic texture, neon effect, 3D effect, or other text effects. \\
Copy Hierarchy Consistency & Strict: yes, no. \\
Copy Appropriateness & Strict: appropriate, inappropriate. \\
Text--Subject Layering & Strict: no overlap, text in front of subject, text behind subject. \\

\midrule
 
\multicolumn{2}{l}{\textit{Background, Scene, and Visual Style}} \\
Background Type & Extensible: indoor home, outdoor nature, urban street, office, industrial or factory, studio, virtual background, none, or other background types. \\
Scene Type & Extensible: solid color, gradient, AI or virtual scene, abstract texture, geometric background, 3D space, none, or other scene types. \\
Background Color Tone & Extensible: warm, cool, neutral, low-saturation, soft, high-contrast, or other color-tone descriptions. \\
Visual Style & Extensible: fresh and natural, colorful, minimalist, cinematic, cyberpunk, premium metallic, matte, gothic, retro, film-grain, or other visual styles. \\
Background--Product Harmony & Extensible: well matched, product highlighted by contrast, neutral, mismatched, or other harmony descriptions. \\
Color Relationship & Extensible: analogous, monochrome, complementary, high contrast, background highlights subject, or other color relationships. \\
Depth of Field & Strict: sharp, slight blur, strong depth-of-field blur, motion blur, none. \\
Lighting Consistency & Extensible: natural lighting, lighting inconsistency, shadow error, missing shadow, floating object, or other lighting issues. \\

\midrule
 
\multicolumn{2}{l}{\textit{Product Presentation}} \\
Product Quantity & Strict: product set, single item, 2--3 items, 4--5 items, large flat-lay assortment. \\
Product Area Ratio & Strict: $>3/4$ dominant, $1/2$--$3/4$ main, $1/4$--$1/2$ moderate, $1/8$--$1/4$ small, $<1/8$ decorative. \\
Product Position & Strict: left, right, center, top, bottom, top-left, top-right, bottom-left, bottom-right, none. \\
Product Orientation & Strict: front view, 45-degree tilt, top-down view, bottom-up view. \\
Product Display State & Strict: static display, in use. \\
Product Completeness & Strict: fully visible, partially cropped, edge only, heavily occluded. \\
Product Overlap & Strict: no overlap, slight overlap, heavy stacking. \\
Product Shadow & Strict: no shadow, natural shadow, reflection. \\

\midrule
 
\multicolumn{2}{l}{\textit{Model and Interaction}} \\
Model Presence & Strict: present, absent. \\
Number of Models & Strict: 1, 2, more than 2, none. \\
Model Type & Extensible: human model, non-human model, virtual or AI model, or other model types. \\
Model Age & Strict: infant, child, young adult, middle-aged, elderly, none. \\
Model Gender & Strict: male, female, neutral, mixed, none. \\
Model Outfit & Extensible: business, casual, sportswear, techwear, retro, ethnic, homewear, pajamas, minimalist, premium, formalwear, eveningwear, none, or other outfit styles. \\
Model Gaze Direction & Strict: looking at camera, looking at product, looking elsewhere, eyes closed or sunglasses, none. \\
Model Action & Extensible: standing, sitting, walking, running, product demonstration, sports or fitness action, facial close-up, none, or other actions. \\
Model--Product Interaction & Extensible: physical contact, gaze interaction, functional demonstration, no interaction, or other interaction types. \\
Inter-Model Physical Interaction & Extensible: no contact, handshake, hug, leaning, side-by-side, object passing, interactive demonstration, none, or other interactions. \\

\midrule
 
\multicolumn{2}{l}{\textit{Logo}} \\
Logo Presence & Strict: present, absent. \\
Logo Position & Strict: left, right, center, top, bottom, top-left, top-right, bottom-left, bottom-right, none. \\
Logo Size & Strict: large, medium, small, none. \\
Logo Count & Strict: 1, 2, 3 or more, none. \\

\end{longtable}
\end{center}

\subsection{A.2 Preliminary Attribute Screening}

Before statistical analysis, we exclude attributes that cannot provide reliable and independent CTR attribution. The main reasons include unstable large-scale extraction, highly subjective labels, severe distribution imbalance, semantic redundancy, and insufficient valid samples. Tab.~\ref{tab:pre_excluded_attributes} summarizes the 14 pre-excluded attributes and their exclusion rationales.

\begin{center}
\small
\setlength{\tabcolsep}{5pt}
\renewcommand{\arraystretch}{1.08}

\begin{longtable}{p{0.25\textwidth}p{0.68\textwidth}}
\caption{Pre-excluded visual attributes and their exclusion rationale.}
\label{tab:pre_excluded_attributes} \\

\toprule
\textbf{Pre-excluded Attribute} & \textbf{Exclusion Rationale} \\
\midrule
\endfirsthead

\toprule
 
\textbf{Pre-excluded Attribute} & \textbf{Exclusion Rationale} \\
\midrule
\endhead

\midrule
\multicolumn{2}{r}{\textit{Continued on next page}} \\
\endfoot

\bottomrule
\endlastfoot

Image Aspect Ratio &
The production scenario enforces square advertising creatives, making aspect ratio uninformative for CTR attribution. \\

Copy Content &
On-image copy is often constrained by platform review, legal requirements, and business rules. Therefore, free-form copy content is excluded as a stable statistical factor, while copy-related visual properties are retained. \\

Copy Alignment &
A single global alignment label is unreliable because different copy blocks within the same creative may follow different alignment patterns. \\

Character Spacing &
Character spacing is visually subtle in most creatives and is difficult to extract consistently at large scale. \\

Font Size &
Font size is highly relative to image scale, resolution, and layout context, making stable automatic annotation difficult across heterogeneous creatives. \\

Font Type &
This attribute substantially overlaps with Font Style \& Effects and may introduce redundant statistical factors. \\

Copy Position &
Copy position is better represented by text--product layout and overall composition rather than by an isolated spatial location label. \\

Copy Appropriateness &
The judgment is highly subjective and lacks a stable objective boundary for large-scale automatic extraction. \\

Logo Position &
Logo-related information is analyzed at a coarser presence level to avoid sparse and unstable subgroups. \\

Logo Size &
Logo size is merged into the coarse logo-presence factor because fine-grained size labels provide limited independent statistical support. \\

Logo Count &
Logo count is merged into logo presence since multiple-logo cases are relatively sparse in the dataset. \\

Inter-Model Physical Interaction &
This attribute only applies to rare multi-model cases and does not provide sufficient samples for robust statistical analysis. \\

Product Overlap &
Product overlap is difficult to label consistently when multiple product sets or dense layouts appear in the same creative. \\

Text--Subject Layering &
The distribution is highly imbalanced because most creatives do not contain explicit text--subject overlap. \\

\end{longtable}
\end{center}

\subsection{A.3 Statistical Analysis and Final Selection}

For the remaining candidate attributes, we combine one-way ANOVA with Random Forest contribution. ANOVA measures the independent association between each attribute and normalized CTR, while the Random Forest captures its contribution under interactions with other attributes. As reported in Tab.~\ref{tab:attribute_statistical_analysis}, most attributes are initially selected if they exhibit either significant ANOVA results ($p<0.05$) or a normalized Random Forest contribution above $5.0\%$. The complete procedure is summarized in Algorithm~\ref{alg:attribute_mining}.

\begin{algorithm}[h]
\caption{CTR-Aware Visual Attribute Mining}
\label{alg:attribute_mining}
\begin{algorithmic}[1]
\REQUIRE Dataset $\mathcal{D}$, initial attribute set $\mathcal{A}_0$, 
$|\mathcal{A}_0|=51$
\ENSURE CTR-sensitive attribute set $\mathcal{A}^{*}$, 
$|\mathcal{A}^{*}|=28$

\FOR{each sample $i\in\mathcal{D}$}
    \STATE $r_i \leftarrow \frac{c_i}{n_i}$
    \STATE $\hat{r}_i \leftarrow \operatorname{Norm}(r_i\mid s_i)$
\ENDFOR

\STATE $\mathcal{A}\leftarrow
\operatorname{PreExclude}(\mathcal{A}_0)$
\STATE $|\mathcal{A}|=37$

\FOR{each $a\in\mathcal{A}$}
    \STATE $\mathcal{G}_a\leftarrow
    \operatorname{Group}(\mathcal{D},a)$
    \STATE $p_a\leftarrow
    \operatorname{ANOVA}(\{\hat{r}_i\}_{i\in\mathcal{G}_a})$
\ENDFOR

\STATE $\mathbf{X}\leftarrow
\operatorname{Encode}(\mathcal{D},\mathcal{A})$
\STATE $f_{\mathrm{RF}}\leftarrow
\operatorname{RFRegress}(\mathbf{X},\hat{\mathbf{r}})$

\FOR{each $a\in\mathcal{A}$}
    \STATE $w_a\leftarrow
    \sum_{j\in\mathcal{B}(a)}
    \operatorname{Imp}_{f_{\mathrm{RF}}}(j)$
\ENDFOR

\STATE $\mathcal{A}^{*}\leftarrow
\operatorname{Select}
\left(
\mathcal{A},
\{p_a\},
\{w_a\},
R_{\mathrm{dist}},
R_{\mathrm{sem}},
R_{\mathrm{edit}}
\right)$

\RETURN $\mathcal{A}^{*}$
\end{algorithmic}
\end{algorithm}

\subsubsection{Analysis of Variance.}

We employ analysis of variance (ANOVA) to examine whether different values of each visual attribute are associated with statistically significant differences in click-through rate (CTR). Since the raw CTR of an image can be influenced by product-level factors, such as product popularity, category, price, and traffic allocation, directly comparing CTR across different products may introduce considerable bias. Therefore, we normalize CTR within each product and use the normalized CTR as the response variable. For each visual attribute, images are grouped according to their corresponding attribute values. ANOVA evaluates whether the mean normalized CTR differs significantly among these groups by decomposing the total variance into between-group and within-group components. The between-group component captures the variation of group means caused by different attribute values, while the within-group component measures the residual variation among samples within the same group. The test statistic is defined as

\begin{equation}
F =
\frac{\mathrm{MS}_{\mathrm{between}}}
{\mathrm{MS}_{\mathrm{within}}},
\end{equation}

where $\mathrm{MS}_{\mathrm{between}}$ and $\mathrm{MS}_{\mathrm{within}}$ denote the mean squares between and within attribute groups, respectively. A larger $F$ value indicates that the differences among attribute groups are relatively larger than the variations within groups. The corresponding significance test evaluates the null hypothesis that all attribute-value groups share the same expected normalized CTR. Rejecting this hypothesis suggests that the attribute has a statistically significant association with relative click performance. Since ANOVA evaluates each attribute independently, it provides an interpretable measurement of the marginal relationship between individual visual attributes and CTR, but does not explicitly capture nonlinear effects or interactions among multiple attributes.

\subsubsection{Random Forest Regression.}

We further employ a random forest regressor to evaluate the contributions of visual attributes under a multivariate setting. Categorical attribute values are first converted into binary indicator features, such that each possible attribute value is represented by an individual feature branch. A random forest is composed of multiple regression trees. Each tree recursively partitions the feature space according to visual attribute values, with the objective of grouping images that have similar normalized CTR values. At each internal node, the model selects a feature and a splitting rule that produce the largest reduction in regression error. In this way, regression trees can model nonlinear relationships and conditional dependencies between visual attributes and CTR without requiring a predefined functional form.

Different regression trees are constructed from randomized samples and feature subsets. Their predictions are aggregated to obtain the final CTR prediction, reducing the instability and overfitting of a single decision tree. The importance of each feature branch is calculated according to the cumulative reduction in regression impurity produced by splits using that branch across all trees. Since one categorical attribute may be expanded into multiple binary branches, we aggregate the importance scores of all branches belonging to the same original attribute:

\begin{equation}
I(a) =
\sum_{j \in \mathcal{B}(a)} I_j,
\end{equation}

where $\mathcal{B}(a)$ denotes the set of binary feature branches derived from attribute $a$, and $I_j$ denotes the importance score of the $j$-th branch. The aggregated importance scores are then normalized across all attributes:

\begin{equation}
\widetilde{I}(a) =
\frac{I(a)}
{\sum_{a' \in \mathcal{A}} I(a')},
\end{equation}

where $\mathcal{A}$ denotes the complete set of visual attributes. Unlike ANOVA, which examines each attribute independently, random forest regression evaluates the predictive contribution of an attribute while considering the effects of other attributes. It can therefore identify attributes that influence CTR through nonlinear patterns or interactions, even when their individual marginal effects are relatively weak. The random forest analysis complements ANOVA by providing a multivariate estimation of the relative contribution of each visual attribute.

\noindent\textbf{Implementation details.}
CTR is normalized within each product to reduce differences in product popularity and user demand. For each candidate attribute, samples are grouped by their extracted values, and one-way ANOVA is applied to evaluate its association with normalized CTR. Categorical attributes are one-hot encoded before training a Random Forest regressor. The importance scores of feature branches belonging to the same attribute are summed to obtain attribute-level contributions. Final selection further considers distribution reliability, semantic redundancy, and practical editability.
\begin{center}
\scriptsize
\setlength{\tabcolsep}{5pt}
\renewcommand{\arraystretch}{1.08}

\begin{longtable}{p{0.48\textwidth}ccc}
\caption{Statistical analysis of candidate visual attributes after preliminary screening.}
\label{tab:attribute_statistical_analysis} \\

\toprule
 
\textbf{Attribute} & \textbf{$p$-value} & \textbf{RF Contr. (\%)} & \textbf{Selected} \\
\midrule
\endfirsthead

\toprule
 
\textbf{Attribute} & \textbf{$p$-value} & \textbf{RF Contr. (\%)} & \textbf{Selected} \\
\midrule
\endhead

\midrule
\multicolumn{4}{r}{\textit{Continued on next page}} \\
\endfoot

\bottomrule
\endlastfoot

Font Style \& Effects & $<0.001$ & 9.45 & Yes \\
Background Color Tone & 0.001 & 8.29 & Yes \\
Copy Density & $<0.001$ & 6.28 & Yes \\
Visual Style & $<0.001$ & 6.08 & Yes \\
Shooting Angle & 0.019 & 6.04 & Yes \\
Composition Symmetry & $<0.001$ & 4.90 & Yes \\
Background--Product Harmony & $<0.001$ & 4.69 & Yes \\
Scene Type & $<0.001$ & 3.82 & Yes \\
Product Orientation & 0.020 & 3.31 & Yes \\
Copy Language & $<0.001$ & 3.07 & Yes \\
Depth of Field & $<0.001$ & 3.01 & Yes \\
Model--Product Interaction & $<0.001$ & 2.96 & Yes \\
Element Density & $<0.001$ & 2.15 & Yes \\
Product Position & $<0.001$ & 2.11 & Yes \\
Model Outfit & $<0.001$ & 2.09 & Yes \\
Copy Hierarchy Consistency & $<0.001$ & 1.97 & No \\
Product Display State & $<0.001$ & 1.89 & Yes \\
Model Type & $<0.001$ & 1.56 & Yes \\
Model Age & $<0.001$ & 1.32 & Yes \\
Model Gaze Direction & $<0.001$ & 1.30 & Yes \\
Model Action & $<0.001$ & 1.22 & Yes \\
Model Gender & $<0.001$ & 1.15 & Yes \\
Text--Product Consistency & $<0.001$ & 0.92 & Yes \\
Product Area Ratio & $<0.001$ & 0.88 & Yes \\
Background Type & $<0.001$ & 0.84 & Yes \\
Number of Models & $<0.001$ & 0.53 & Yes \\
Model Presence & $<0.001$ & 0.14 & Yes \\
Product Quantity & 0.275 & 5.94 & Yes \\
Logo Presence & 0.033 & 2.55 & Yes \\
Color Relationship & 0.920 & 3.17 & No \\
Product Shadow & $<0.001$ & 2.26 & No \\
Number of Visual Centers & 0.345 & 1.74 & No \\
Visual Layering & $<0.001$ & 0.90 & No \\
Element Alignment & 0.923 & 0.57 & No \\
Lighting Consistency & 0.642 & 0.51 & No \\
Product Completeness & 0.178 & 0.35 & No \\
Visual Balance & 0.082 & 0.06 & No \\

\end{longtable}
\end{center}

\noindent\textbf{Selection exceptions.}
Most attributes follow the joint criterion of ANOVA significance and Random Forest contribution. A few attributes are manually adjusted according to semantic redundancy, distribution reliability, and practical editability. Product Quantity is retained despite its weak ANOVA significance because it has a high Random Forest contribution and directly corresponds to actionable product-presentation edits. Copy Hierarchy Consistency, Product Shadow, and Visual Layering are removed despite significant ANOVA results because they are either redundant with retained attributes or dominated by unstable value distributions. Color Relationship is removed because its ANOVA significance is weak and its semantics largely overlap with Background Color Tone and Background--Product Harmony. Product Completeness and Visual Balance are also removed due to weak model contribution and redundancy with Product Area Ratio, Product Display State, Composition Symmetry, and Element Density.

\clearpage

\section{B. Instruction-Tuning Dataset Examples}

\begin{figure}[!htbp]
\centering
\includegraphics[
    width=0.99\textwidth,
    height=0.78\textheight,
    keepaspectratio
]{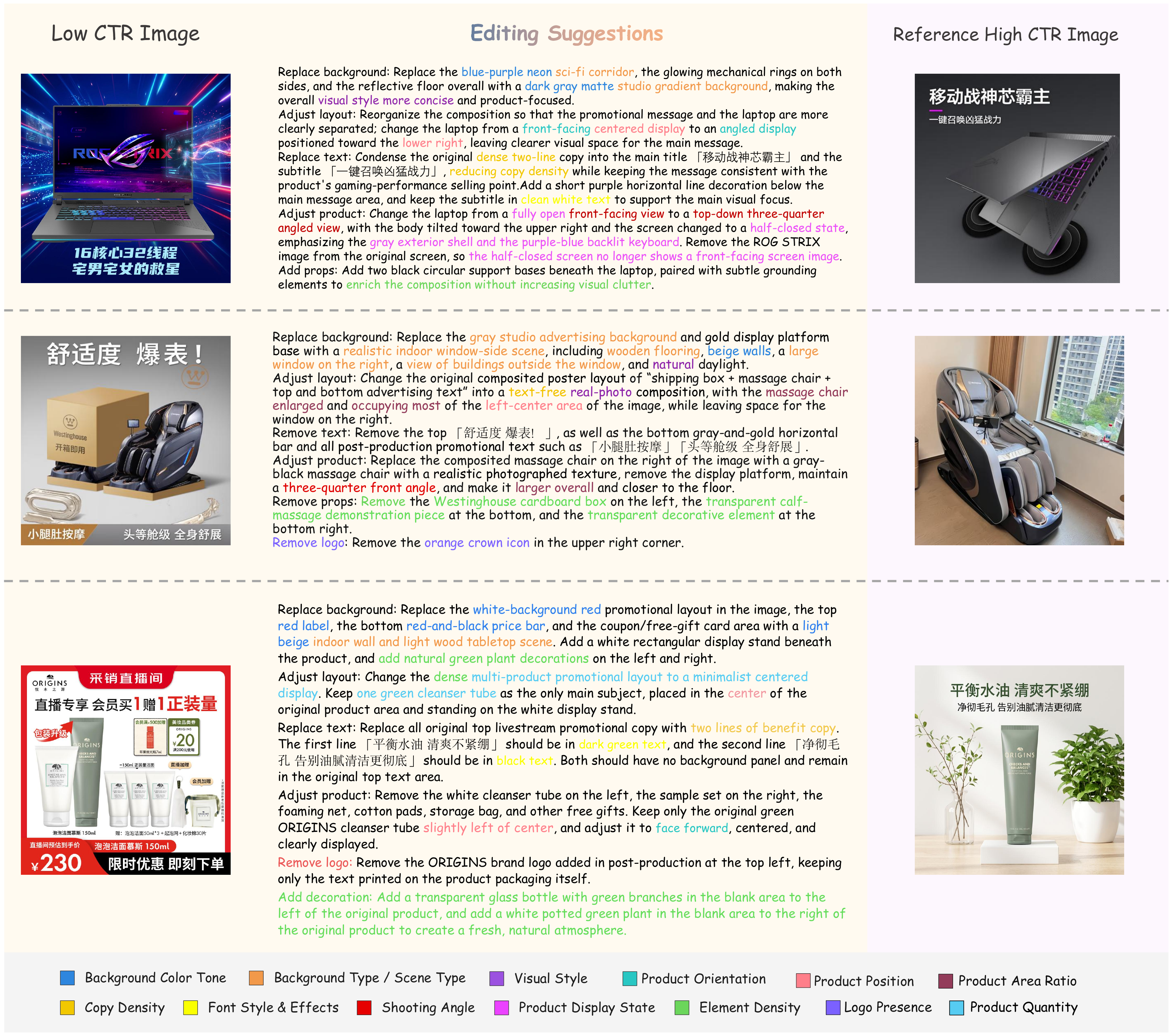}
\caption{Examples of the instruction-tuning dataset. Each row shows the low-CTR input creative, its target editing suggestion, and the same-product high-CTR reference used only during data construction. Colored phrases indicate the involved CTR-sensitive attributes.}
\label{fig:dataset_examples}
\end{figure}

Fig.~\ref{fig:dataset_examples} presents representative instruction-tuning samples. Each sample contains a low-CTR creative as the model input and its corresponding editing suggestion as the supervision target. The higher-CTR image of the same product is used only to construct and validate the editing suggestion and is not provided to the model during training. Colored phrases indicate the CTR-sensitive attributes involved in each editing operation.

\clearpage
\section{C. Additional Qualitative Results}

\begin{figure}[!htbp]
\centering
\includegraphics[
    width=0.99\textwidth,
    height=0.85\textheight,
    keepaspectratio
]{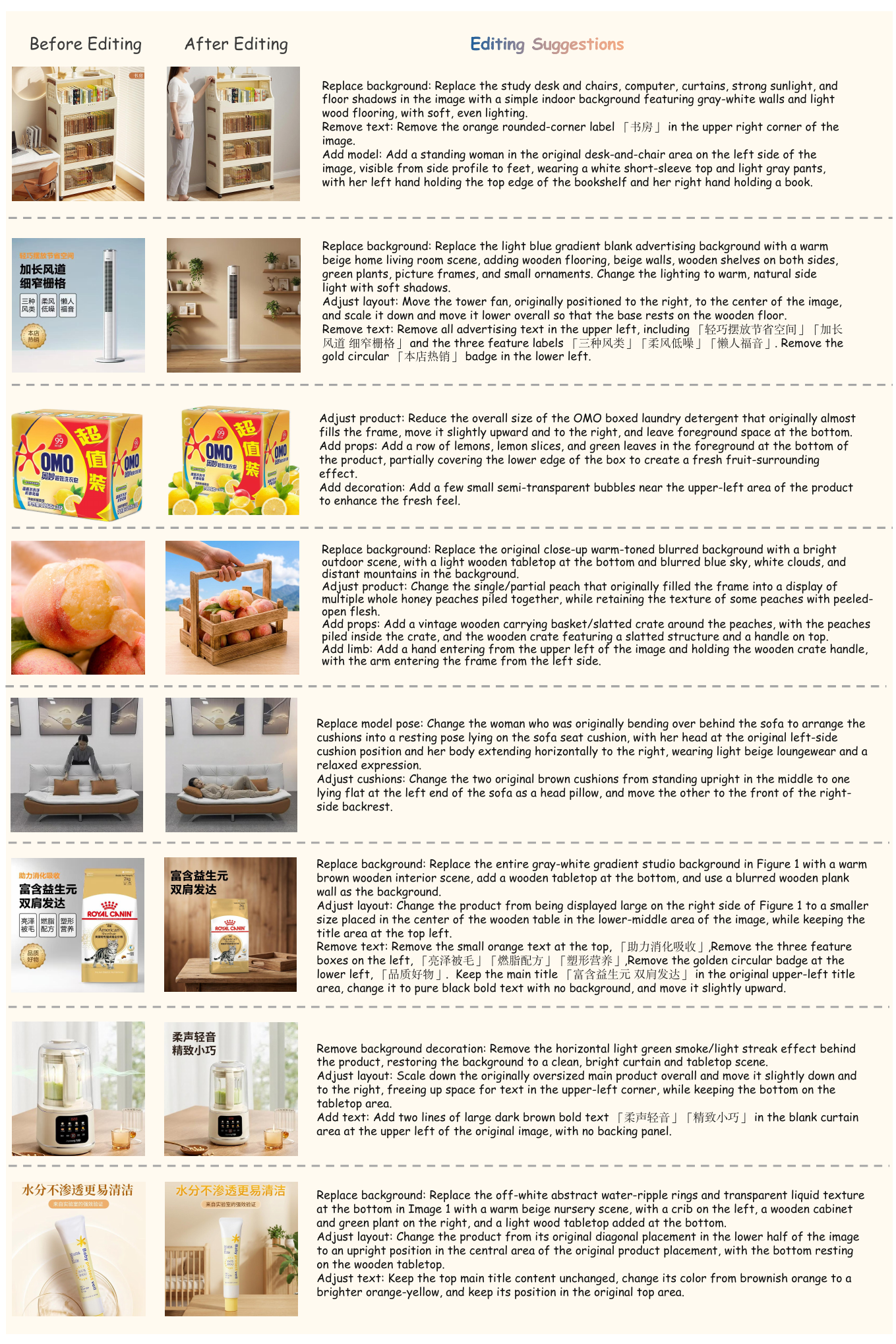}
\caption{Qualitative examples of CTR-driven ad image modification.}
\label{fig:qualitative_results}
\end{figure}

Fig.~\ref{fig:qualitative_results} presents additional examples of CTR-driven ad image editing. ECC generates executable suggestions covering background, layout, product presentation, visual style, and copywriting. The resulting creatives show clearer visual organization and improved online CTR, while preserving the identity and core content of the advertised products.

\end{document}